\documentclass[runningheads]{llncs}
\usepackage{graphicx}
\usepackage{booktabs}
\usepackage{multirow}
\usepackage[normalem]{ulem}
\usepackage[sectionbib, numbers]{natbib}
\useunder{\uline}{\ul}{}
\usepackage{caption}
\usepackage{amsmath}
\usepackage{framed}
\usepackage{mdframed}
\usepackage{lipsum}
\usepackage{hyperref}
\usepackage{subcaption}
\usepackage{footnote}
\usepackage{enumitem}
\usepackage{longtable}
\usepackage{color}
\usepackage[table,dvipsnames]{xcolor}
\usepackage{adjustbox}
\usepackage{marvosym}
\usepackage{diagbox}
\newcolumntype{M}[1]{>{\centering\arraybackslash}m{#1}}

\definecolor{effort}{HTML}{6DFCFF}
\definecolor{outcome}{HTML}{F4CCCC}
\DeclareCaptionLabelSeparator{custom}{. }
\begin{document}

\title{Comparative Analysis of GPT-4 and Human Graders in Evaluating Praise Given to Students in Synthetic Dialogues}

% \titlerunning{Automatic Explanatory Feedback to Human Tutors}

\author{Dollaya Hirunyasiri \and
Danielle R. Thomas\and
Jionghao Lin\and
Kenneth R. Koedinger\and
Vincent Aleven
}
\authorrunning{Dollaya \textit{et al.}}
% If there are more than two authors, '\textit{et al.}' is used.
%
\institute{Carnegie Mellon University, Pittsburgh, PA, USA\\
\email{dhirunya@andrew.cmu.edu}\\
\email{\{Drthomas,Jionghao,Koedinger\}@cmu.edu}\\
\email{aleven@cs.cmu.edu}
}
\maketitle              % typeset the header of the contribution
\vspace{-3mm}
\begin{abstract}
Research suggests that providing specific and timely feedback to human tutors enhances their performance. However, it presents challenges due to the time-consuming nature of assessing tutor performance by human evaluators. Large language models, such as the AI-chatbot ChatGPT, hold potential for offering constructive feedback to tutors in practical settings. Nevertheless, the accuracy of AI-generated feedback remains uncertain, with scant research investigating the ability of models like ChatGPT to deliver effective feedback. In this work-in-progress, we evaluate 30 dialogues generated by GPT-4 in a tutor-student setting. We use two different prompting approaches, the zero-shot chain of thought and the few-shot chain of thought, to identify specific components of effective praise based on five criteria. These approaches are then compared to the results of human graders for accuracy. Our goal is to assess the extent to which GPT-4 can accurately identify each praise criterion. We found that both zero-shot and few-shot chain of thought approaches yield comparable results. GPT-4 performs moderately well in identifying instances when the tutor offers specific and immediate praise. However, GPT-4 underperforms in identifying the tutor's ability to deliver sincere praise, particularly in the zero-shot prompting scenario where examples of sincere tutor praise statements were not provided. Future work will focus on enhancing prompt engineering, developing a more general tutoring rubric, and evaluating our method using real-life tutoring dialogues.


\end{abstract}
\section{Introduction}

Tutoring is among the most highly personalized and consistently impactful interventions known to improve student learning \cite{kraft2021blueprint, lin2022good}. Despite the known positive impacts of tutoring on student learning, there is a known shortage of trained tutors, with many available tutors lacking experience and the necessary competency skills to be successful in the field \cite{thomas2023tutor}. In recent years, although tutor training programs have been developed, most do not provide tutors with specific formative feedback during training, and little research exists on tutors receiving specific feedback on their actual tutoring practices. Recent advances in pre-trained large language models, such as the well-known AI-chatbot ChatGPT, have made it possible to provide specific and explanatory feedback to learners \cite{chen2022program}. We propose that the use of large language models to provide tutors with effective formative feedback on their actual tutoring is a promising use case.

The ability of GPT-4 to accurately evaluate components of praise given to students, which can be determined by comparing it to human expert evaluations, is a critical component of providing effective feedback, and as such, serves as our starting point. Moreover, the accuracy of AI-generated tutor feedback for the purpose of improving tutor learning and performance has not been well researched, if at all. In this work-in-progress, we used simulated dialogues to assess the capability of GPT-4 in providing accurate feedback to human tutors regarding their delivery of effective praise to students. To this end, the primary research question addressed is:

\begin{itemize}[leftmargin=.42in]
    \item[\textbf{RQ: }] Can GPT-4 accurately assess components of effective human tutor's praise to students and, in particular, what is the comparative accuracy between zero-shot and few-shot chain of thought prompting approaches?
\end{itemize}

\section{Related Work}
\subsection{High-Quality Feedback}
Feedback is one of the most powerful influences on student achievement and can significantly impact learning outcomes and performance \cite{ryan2021designing, hattie2007power, lin2023learner}. Effective feedback is described as having many characteristics, particularly: 1) being targeted, linked to specific goals and learning objectives; 2) being progress-oriented and constructive, focusing on the learning process and supporting a growth mindset; 3) being timely, as providing immediate and frequent feedback often benefits students' academic performance \cite{ryan2021designing, goodwin2012good, demszky2021can}. However, providing learners with timely, explanatory feedback, or in this case, offering timely feedback to online tutors while they are actively tutoring students is laborious and expensive when using human evaluators \cite{dai2023can}. To facilitate the feedback provision process, Demszky \textit{et al.} \cite{demszky2021can} provide automated, individualized feedback to over one thousand instructors on their teaching sessions within an online computer course. Instructors received the feedback via email within 2-4 days. This automatic, formative feedback tool improved instructors' uptake of student contributions by 27\%, with preliminary evidence suggesting it also increased students' satisfaction with assignments and the course itself \cite{demszky2021can}. These promising findings underscore the potential that more timely feedback—either occurring in real time or shortly after—a tutoring session could enhance student contribution and performance. Despite the known positive impact of feedback on educators' performance and the global interest in leveraging large language models (LLMs) for communicative tasks, there is currently a lack of research on the use of LLMs for generating feedback on tutoring. 

\subsection{Tutoring Competencies \& Giving Effective Praise}
There is limited research on the key competencies and components of effective tutoring \cite{chhabra2022evaluation}, with many qualities of impactful tutoring challenging to measure or assess (e.g., building a relationship with the student) in practice. The National Student Support Accelerator (2021), a think tank emanating from the Annenberg Institute at Brown University that focuses on disseminating research and advancing developments in tutoring, has created a rubric for evaluating the effectiveness of tutors in facilitating sessions. The rubric contains three main criteria for assessing a tutoring session: 1) The tutor effectively employs tutoring facilitation strategies; 2) The tutor identifies and addresses potential student misconceptions or confusions; and 3) The tutor explains content clearly and correctly. Each criterion is measured on a 1-5 Likert-like scale, from \textit{``lacking''} to \textit{``exemplary''}, respectively \cite{NSSA2021}.

Our recent research, surveying 18 partnering members across several tutoring organizations, determined that the most important perceived tutoring skills were the ability to engage and motivate students and build successful relationships with them \cite{chhabra2022evaluation}. From this research, we developed a super competency framework called SMART, standing for \underline{S}ocial-emotional learning, \underline{M}astering content, \underline{A}dvocacy, \underline{R}elationships, and \underline{T}echnology. Mastering Content, which pertains to a tutor's ability to comprehend mathematical pedagogy and apply effective tutoring skills, was identified as a crucial element of effective tutoring. Within this dimension, there are multiple scenario-based lessons covering a range of content. We selected the lesson titled \textit{Giving Effective Praise} as our starting point, given its critical role in fostering and maintaining student motivation and engagement. The lesson objectives from \textit{Giving Effective Praise} state that upon completion of the lesson, tutors will be able to: explain how to increase student motivation by giving effective praise; identify features of effective praise; and apply strategies by responding to students through praise \cite{thomas2023tutor}. Tutors should strive to praise students for their effort, acknowledging the learning process, and not necessarily the outcome, such as getting the problem correct \cite{dweck2006mindset}. The five key criteria for productive, process-focused praise used as a rubric in this work state that praise is: 1) sincere, earned, and truthful; 2) specific by giving details of what the student did well; 3) immediate, with praise given right after the student’s action; 4) authentic, not repeated often; and 5) focused on the learning process, not ability \cite{thomas2023tutor}.

Given the known importance of effective praise on student motivation and performance, can large language models like GPT-4 pick up on the use of these strategies when analyzing tutor-student interaction data (i.e., tutor-student chat logs or transcripts)? If so, this would open the door to using large language models, such as GPT-4, to generate timely, impactful, and formative feedback to tutors during their actual tutoring sessions.

\subsection{Using Large Language Models to Give Feedback}
Large language models (LLMs) are trained using deep learning to produce text that resembles human writing. Trained on a vast array of sources, such as Wikipedia, webpages, written materials, and practically anything curated on the internet, the text generated by neural LLMs often mirrors the written language of most humans. We focus on ChatGPT using GPT-4, a general pre-trained large multimodal model capable of accepting both image and text inputs. OpenAI \cite{openai2023gpt4} asserts, \textit{``while less capable than humans in real-world scenarios, [GPT-4] exhibits human-level performance on various professional and academic benchmarks.''} This current investigation seeks to determine if identifying tutors’ ability to give effective praise to students is an academic benchmark within GPT-4's capabilities.

The application of LLMs to provide feedback is a growing research area within education \cite{dai2023can}, with researchers striving to identify the limits of these models' capabilities. The use of LLMs to provide direct feedback to students, rather than tutors, has been explored by many researchers using various pre-trained models. For example, Jai \textit{et al.} \cite{jia2022insta} used BART and found that AI-generated feedback was near-human in performance, while Li and Xing \cite{li2021natural}, employing GPT-based models, concluded that providing emotional support via contextual replies to learners in massive open online courses (MOOCs) was comparable to humans. In a study more closely aligned with our current work, Dai \textit{et al.} \cite{dai2023can} demonstrated that ChatGPT was more capable than human instructors at genera  -ting detailed feedback that fluently summarizes students’ performance. Despite these promising findings involving LLM’s ability to provide feedback to students, there exists very little research on its application to tutor feedback. Thomas \textit{et al.} \cite{thomas2023aied} leveraged ChatGPT to generate synthetic tutor responses from real-life tutoring scenarios within the previously discussed lesson, \textit{Giving Effective Praise}. Thomas \textit{et al.} \cite{thomas2023aied} found that human-created training sets outperformed AI-generated training sets for creating automated short answer grading systems, with ChatGPT-generated tutor responses often lacking the nuance and variety evident within human-sourced tutor responses. Nevertheless, leveraging ChatGPT to evaluate human tutors' effectiveness in giving praise to students represents an interesting and novel use case.

\vspace{-3mm}

\subsection{Prompt Engineering}
Prompt engineering, also known as in-context prompting, is the strategic creation and fine-tuning of prompts aimed at guiding a language model's behavior to yield specific outcomes. This process is achieved without the necessity of modifying the model's inherent architecture or parameters. As an empirical field, prompt engineering necessitates extensive experimentation and testing, considering the variations in the outcomes generated by identical prompts across different models \cite{wei2022chain}. Chain-of-Thought (CoT) prompting is a technique that breaks down complex, multi-step problems into more manageable, intermediate steps. This process aids language models in following a logical sequence, where each subsequent prompt builds upon the prior one, thus stimulating reasoning. Within the context of CoT prompting, two key methodologies exist: zero-shot and few-shot prompting. Zero-shot CoT prompting is a standalone approach that relies solely on the instructions embedded within the prompt. Conversely, few-shot CoT prompting incorporates examples to instruct the model on generating appropriate responses. Zero-shot and few-shot prompting are two fundamental approaches often championed in numerous large language model (LLM) studies, commonly employed for benchmarking LLM performance \cite{wei2022chain, li2023can}.

\section{Method}
\subsection{Creation of Synthetic Tutoring Dialogues}
Due to the limited availability of real-life tutor-student dialogues, we used synthetic chat logs generated by prompting GPT-4. While we acknowledge the necessity of validating our findings with real-life dialogues, the current study is useful as a proof of concept and serves as a simulation or model, pending access to real-life tutor-student dialogues. We used GPT-4 to generate 30 synthetic tutor-student dialogues. Among these dialogues, the average number of words per dialogue was 253 \textit{(SD = 45.0)}; the tutor's words per dialogue averaged 180 \textit{(SD = 38.6)}; and the student's words per dialogue averaged 56.8 \textit{(SD = 23.7)}. Due to the limited space, we attached other prompting strategies and synthetic tutoring dialogues in the digital appendix\footnote{\url{https://github.com/DollayaDollayaDollaya/AIEDWorkshop}}. An example of a tutor-student dialogue generated by GPT-4 is shown in the Example 1:

    \vspace{-3mm}
\captionsetup[table]{skip=0pt,singlelinecheck=off, labelsep=custom, name=\textbf{Example},skip=5pt,labelfont={small}, font= small}
\begin{table}[ht]
    \centering
    \caption{An example of synthetic tutor-student dialogue generated by GPT-4}
    \resizebox{0.9\textwidth}{!}{%
\renewcommand{\arraystretch}{1.25}
    \begin{tabular}{p{12cm}}
    % \hline
    \rule{-3pt}{10pt}
    \textbf{Tutor:} \textit{Good evening! Let's start with this problem. ``Last week 24,000 fans attended a football match. Three times as many bought tickets this week, but one-sixth of them canceled their tickets. How many are attending this week?''} \\
    \textbf{Student:} \textit{I think we need to first calculate three times the fans from last week, right?} \\
    \textbf{Tutor:} \textit{Absolutely, you're on the right track! Now, how much does that make?} \\
    \textbf{Student:} \textit{That would be 72,000.} \\
    \textbf{Tutor:} \textit{Excellent multiplication! Yes, you're correct. Now, remember one-sixth of them canceled their tickets. What should we do next?} \\
    \textbf{Student:} \textit{We need to subtract one-sixth of 72,000 from 72,000.} \\
    \textbf{Tutor:} \textit{That's right! I see you've clearly grasped the concept of fractions. Could you calculate it for me?} \\
    \textbf{Student:} \textit{The answer should be 60,000.} \\
    % \hline
    \end{tabular}
    }
    \vspace{-6mm}
\end{table}
% \vspace{-3mm}
\subsection{Human Grader Identification of Praise Criteria  }
To evaluate the accuracy of GPT-4, we initially recruited three human graders, each with over five years of teaching experience. These graders were tasked with identifying effective praise within synthetic tutoring dialogues. Before beginning this task, they each completed a lesson titled  \textit{Giving Effective Praise}. This lesson clearly defines effective praise and trains learners on how to apply it. Additionally, the human graders were provided with a rubric that includes five criteria for identifying the different aspects of praise. This rubric, proposed by \cite{thomas2023tutor} (introduced in Section 2.2), includes five key criteria and their notation (in parenthesis) are, as follows:  Praise is: 1) sincere, earned, and truthful \textit{(Sincere)}; 2) specific by giving details of what the student did well \textit{(Specific)}; 3) immediate, with praise given rights after the student’s action \textit{(Immediate)}; 4) authentic, not repeated often \textit{(Authentic)}; and 5) focused on the learning process, not ability \textit{(Process-focused)}. To arrive at the final grading for each dialogue, we used majority voting among the human graders. For instance, if two or more graders assessed that a particular chat log did not meet criterion 1 \textit{(Sincere)}, we followed their consensus and regarded that as the ground truth. Finally, we employed Fleiss' Kappa \cite{fleiss1971measuring} to measure the inter-rater reliability among the three human graders (shown in Table 1).
\vspace{-6mm}
\setcounter{table}{0}
\captionsetup[table]{skip=0pt,singlelinecheck=off, labelsep=custom, name=\textbf{Table},skip=5pt,labelfont={small}, font= small}

\begin{table*}[!htb]
\centering
\caption{Agreement among three human graders on identifying praise criteria proposed by \cite{thomas2023tutor}.}
\resizebox{1\textwidth}{!}{%
\renewcommand{\arraystretch}{1.5}
\begin{tabular}{m{.25\textwidth}M{.25\textwidth}M{.25\textwidth}M{.25\textwidth}}
\hline
\textbf{Praise Criteria} & \textbf{Agreement score} & \textbf{Fleiss' Kappa} & \textbf{Interpretation} \\ \hline
1-\textit{Sincere}                & 84.44\%                   & 0.60                   & \textit{Moderate}       \\
2-\textit{Specific}               & 73.33\%                   & 0.44                   & \textit{Moderate}       \\
3-\textit{Immediate}              & 68.89\%                   & 0.34                   & \textit{Fair}           \\
4-\textit{Authentic}              & 88.89\%                   & 0.69                   & \textit{Substantial}    \\
5-\textit{Process-focused}        & 64.44\%                   & 0.29                   & \textit{Fair}           \\ \hline
\end{tabular}
}
\vspace{-6mm}
\end{table*}
% \vspace{-6mm}

\subsection{Prompting GPT-4 to Identify Praise Criteria}
We prompted GPT-4 to identify instances of praise in the dialogues based on the specific criteria provided. Recognizing that the effectiveness of GPT-4 is largely influenced by the prompt engineering strategies used, we implemented two approaches: zero-shot and few-shot Chain of Thought (CoT) prompting. This generated two sets of results. These results were then compared to the assessments made by human graders, using precision, recall, and F1 scores as metrics. Due to space constraints, we have included the zero-shot CoT and few-shot CoT prompts in the \href{https://github.com/DollayaDollayaDollaya/AIEDWorkshop}{digital appendix}.

\section{Results}
\subsection{Comparison of GPT-4 and Human Grader Performance}
We compared the results from GPT-4, using both zero-shot CoT and few-shot CoT prompting, with the consensus results from the human graders. The results are presented in Table 2. Both the zero-shot CoT and few-shot CoT approaches performed well in detecting elements of \textit{specific} praise (i.e., detailing what the student did well) and \textit{immediate} praise (i.e., given right after the student's action). We posit that the relative straightforwardness and clear nature (i.e., the tutor either delivered praise immediately after the student's action or they did not) of criterion 2 and 3, \textit{specific} and \textit{immediate} praise respectively, make them easier to detect by GPT-4 and human graders when present, compared to the remaining criteria. Both the zero-shot and few-shot CoT prompting methods for detecting \textit{specific} praise had the lowest performance comparison between GPT-4 and the human graders, with F1 scores of 0.54 and 0.67, respectively.

\vspace{-3mm}
% \usepackage{multirow}
\begin{table*}[!htb]
\centering
\caption{The comparison of the performance of GPT-4 and the consensus of human graders using both zero-shot and few-shot CoT prompting methods, as illustrated through precision, recall, and F1 scores by praise criteria, demonstrated good performance in detecting specific and immediate praise criteria.}
\resizebox{\textwidth}{!}{%
\renewcommand{\arraystretch}{1.5}
\begin{tabular}{m{.22\textwidth}M{.15\textwidth}M{.15\textwidth}M{.15\textwidth}M{.15\textwidth}M{.15\textwidth}M{.15\textwidth}}
\hline
\multirow{2}{*}{\textbf{Praise Criteria}} & \multicolumn{3}{c}{\textbf{Zero-shot CoT}} & \multicolumn{3}{c}{\textbf{Few-shot CoT}} \\ \cline{2-7} 
                                          & \textbf{Precision}       & \textbf{Recall}       & \textbf{F1 score}         & \textbf{Precision}       & \textbf{Recall}       & \textbf{F1 score}        \\ \hline
1-\textit{Sincere}                                 & 0.37            & 1.00         & 0.54       & 0.50            & 1.00         & 0.67      \\
2-\textit{Specific}                                & 0.75            & 0.92         & 0.83       & 0.85            & 0.85         & 0.85      \\
3-\textit{Immediate}                               & 0.75            & 0.90         & 0.82       & 0.72            & 0.90         & 0.80      \\
4-\textit{Authentic}                               & 0.60            & 1.00         & 0.75       & 0.63            & 0.83         & 0.71      \\
5-\textit{Process-focused}                         & 1.00            & 0.50         & 0.67       & 1.00            & 0.50         & 0.67      \\ \hline
\end{tabular}
}
\vspace{-6mm}
\end{table*}
\vspace{-3mm}

\subsection{Comparison of Zero-shot and Few-shot Prompting}
The performance of zero-shot and few-shot CoT prompting methods showed a significant degree of similarity. To quantitatively assess the inter-rater agreement between these two approaches, we utilized Cohen's kappa statistical measure. The analysis in Table 3 showed a substantial level of agreement between the zero-shot and few-shot CoT prompting techniques, suggesting a strong consistency in their performance. Specifically, there was \textit{nearly perfect} agreement (93.33\%) in identifying \textit{authentic} and \textit{process-focused} praise criteria, with substantial agreement in recognizing \textit{sincere} and  \textit{specific} praise.

\vspace{-3mm}
\begin{table*}[!htb]
\centering
\caption{Inter-rater reliability between zero-shot and few-shot CoT prompting methods. Notice \textit{near perfect} agreement for the detection of \textit{authentic} and \textit{process-focused} praise criteria.
}
\resizebox{1\textwidth}{!}{%
\renewcommand{\arraystretch}{1.5}
\begin{tabular}{m{.25\textwidth}M{.25\textwidth}M{.25\textwidth}M{.25\textwidth}}
\toprule
\textbf{Praise Criteria} & \textbf{Agreement score} & \textbf{Cohen's Kappa} & \textbf{Interpretation} \\ \midrule
1-\textit{Sincere}                & 83.33\%          & 0.66          & \textit{Substantial}    \\
2-\textit{Specific}               & 90.00\%          & 0.80          & \textit{Substantial}    \\
3-\textit{Immediate}              & 83.33\%          & 0.44          & \textit{Moderate}       \\
4-\textit{Authentic }             & 93.33\%          & 0.84          & \textit{Near perfect}   \\
5-\textit{Process-focused}        & 93.33\%          & 0.85          & \textit{Near perfect}    \\ \bottomrule
\end{tabular}
}
\vspace{-3mm}
\end{table*}

\subsection{Strengths and Weaknesses of GPT-4 Across Praise Criteria}
In reference to Table 2, it's evident that GPT-4 performed well in identifying specific and immediate types of praise, as indicated by the favorable performance measures (F1 $>$80\%). Examples 2, 3, and 4 illustrate selected tutor dialogues and responses generated by GPT-4 using few-shot CoT prompting that align with the majority decision of the human graders. It's worth noting that we chose to highlight criteria 2 (\textit{Specific}), 3 (\textit{Immediate}), and 4 (\textit{Authentic}), which have relatively high F1 score.

\vspace{-6mm}

\setcounter{table}{1}
\captionsetup[table]{skip=0pt,singlelinecheck=off, labelsep=custom, name=\textbf{Example},skip=5pt,labelfont={small}, font= small}
\begin{table}[ht]
    \centering
    \caption{Does not meet criterion 2 \textit{(specific)}}
    \resizebox{0.9\textwidth}{!}{%
\renewcommand{\arraystretch}{1.25}
    \begin{tabular}{p{12cm}}

    \rule{-3pt}{10pt}
    \textbf{Tutor:} \textit{What a superstar! Your intellectual prowess is unmatched.} \\
    \textbf{GPT-4’s rationale:} The praise is overly broad and doesn’t give specific details about what the student did well. The student merely agreed with the tutor [dialogue not shown], and the tutor’s response is disproportionate. \\

    \end{tabular}
    }
    \vspace{-3mm}
\end{table}

\vspace{-12mm}

\captionsetup[table]{skip=0pt,singlelinecheck=off, labelsep=custom, name=\textbf{Example},skip=5pt,labelfont={small}, font= small}
\begin{table}[ht]
    \centering
    \caption{Meets criterion 3 \textit{(immediate)}}
    \resizebox{0.9\textwidth}{!}{%
\renewcommand{\arraystretch}{1.25}
    \begin{tabular}{p{12cm}}
    % \hline
    \rule{-3pt}{10pt}
    \textbf{Student:} \textit{Oh, I see. I should have multiplied it by 100, not 50. I'll  remember that next time.} \\
    \textbf{Tutor:} \textit{That's the spirit! I'm really impressed with your enthusiasm to learn from your mistakes. Keep it up!} \\
    \textbf{GPT-4’s rationale:} The tutor praises the student's willingness to learn immediately after the student admits their mistake and expresses an intention to remember the correct method for the future. \\
    % \hline
    \end{tabular}
    }
    \vspace{-3mm}
\end{table}
\vspace{-12mm}

\captionsetup[table]{skip=0pt,singlelinecheck=off, labelsep=custom, name=\textbf{Example},skip=5pt,labelfont={small}, font= small}
\begin{table}[ht]
    \centering
    \caption{Does not meet criterion 4 \textit{(authentic)}}
    \resizebox{0.9\textwidth}{!}{%
\renewcommand{\arraystretch}{1.25}
    \begin{tabular}{p{12cm}}
    % \hline
    \rule{-3pt}{10pt}
    \textbf{Tutor:} \textit{Oh, you are such a genius! You are so fast and responsive. Your ability to pick an answer quickly is absolutely admirable.} \\
    \textbf{[Tutor and student talking.]} \\
    \textbf{Student:} \textit{...But… I just agreed with you.} \\
    \textbf{Tutor:} \textit{That’s right! You agreed correctly…} \\
    \textbf{Student:} \textit{Multiply?} \\
    \textbf{Tutor:} \textit{Brilliant! Absolutely brilliant! Your mathematical mind is truly something to behold. You are a natural-born mathematician. It's amazing!} \\
    \textbf{Student:} \textit{I’m just guessing} \\
    \textbf{Tutor:} \textit{And what an excellent guess it was!... } \\
    \textbf{GPT-4’s rationale:} The tutor repeats similar forms of praise multiple times, making the praise feel less authentic and repetitive. \\
    % \hline
    \end{tabular}
    }
    \vspace{-3mm}
\end{table}
\vspace{-3mm}

Then, we examined instances where GPT-4 disagreed with the majority of human graders, underperforming in its ability to detect different praise criteria. We focused on criteria 1 (\textit{sincere}) and 5 (\textit{process-focused}), for which GPT-4 received F1 score of 0.67, lower than the other criteria. In Example 5 below, after the student provided three incorrect responses before eventually arriving at the correct answer, human graders interpreted the subsequent praise as insincere (criteria 1), contending that the student's achievement didn't entirely warrant the commendation. In contrast, GPT-4 failed to incorporate this context into its evaluation. It seemingly focused solely on the immediate conversation, noting that the student had provided a correct answer, and concluded that the praise was therefore sincere and deserved. In Example 6, GPT-4 misinterpreted the tutor's praise for the student's efforts. The tutor's compliment, i.e., \textit{``You're showing a keen ability to recollect and apply important mathematical principles,''} was interpreted by GPT-4 as praise for ability, due to the inclusion of the term \textit{``ability''}. However, human graders perceived this compliment as being directed towards the learning process. In this regard, GPT-4's interpretation deviated from the human graders' consensus.

\vspace{-6mm}

\captionsetup[table]{skip=0pt,singlelinecheck=off, labelsep=custom, name=\textbf{Example},skip=5pt,labelfont={small}, font= small}
\begin{table}[ht]
    \centering
    \caption{Human-grader majority says it does not meet criterion 1 \textit{(sincere)}, but GPT-4 says it does meet}
    \resizebox{0.9\textwidth}{!}{%
\renewcommand{\arraystretch}{1.25}
    \begin{tabular}{p{12cm}}
    % \hline
    \rule{-3pt}{10pt}
    \textbf{Student:} \textit{Ok. I think it is... A?} \\
    \textbf{Tutor:} \textit{Good start! (...) However, the result doesn't match the expected value.} \\
    \textbf{Student:} \textit{Ok, I'll try again. Maybe it's D?} \\
    \textbf{Tutor:} \textit{(...) Unfortunately, it's not equation D either. But, your perseverance is really commendable.} \\
    \textbf{Student:} \textit{I see. What about C?} \\
    \textbf{Tutor:} \textit{That's a good guess too. Unfortunately, it's not equation C either. However, your willingness to keep guessing and learning from your mistakes is extremely praiseworthy.} \\
    \textbf{Student:} \textit{Oh... okay. So it must be B then.} \\
    \textbf{GPT-4’s rationale:} The tutor sincerely appreciates the student's effort and perseverance. The praise is based on the student's actual actions. \\
    % \hline
    \end{tabular}
    }
    \vspace{-3mm}
\end{table}
\vspace{-9mm}

\captionsetup[table]{skip=0pt,singlelinecheck=off, labelsep=custom, name=\textbf{Example},skip=5pt,labelfont={small}, font= small}
\begin{table}[ht]
    \centering
    \caption{Human-grader majority says it meets criterion 5 \textit{(process-focused)} but GPT-4 says it does not meet}
    \resizebox{0.9\textwidth}{!}{%
\renewcommand{\arraystretch}{1.25}
    \begin{tabular}{p{12cm}}
    % \hline
    \rule{-3pt}{10pt}
    \textbf{Tutor:} \textit{You're showing a keen ability to recollect and apply important mathematical principles.} \\
    \textbf{GPT-4’s rationale:} The tutor’s praise focuses on the student’s ability, rather than specifically praising their learning process. \\
    % \hline
    \end{tabular}
    }
    \vspace{-3mm}
\end{table}

\vspace{-6mm}

\section{Discussion}
\textbf{GPT-4 exhibited proficiency in detecting specific and immediate praise, but it struggled to recognize sincerity.} We hypothesize that GPT-4's superior performance in detecting \textit{specific} and \textit{immediate} praise is due to the relatively straightforward criteria for these types, while assessing sincerity in praise statements demands more nuanced judgment and perhaps a greater level of social-affective understanding (e.g., politeness \cite{lin2022exploring, lin2023role}), which human graders possess. We noticed that it was particularly challenging for GPT-4 to identify sincerity, especially during the zero-shot CoT prompting. By including nuanced and varied examples of tutor praise statements, deemed sincere by human graders, in few-shot prompting strategies, we might enhance GPT-4's performance in recognizing this type of praise.

Both zero-shot and few-shot CoT prompting exhibited comparable performance. Zero-shot and few-shot learning methods demonstrated similar results, with both falling short in detecting sincerity in praise (with F1 scores of 0.54 and 0.67, respectively) compared to their performance on other praise criteria. Various techniques for fine-tuning language models exist, particularly for zero-shot learning, such as instruction tuning \cite{wei2021finetuned}. Therefore, further research into enhancing zero-shot and few-shot learning methods is necessary to improve the performance of both models.

\vspace{-3mm}

\subsection{Limitations}

The current study has several limitations. First, the lack of availability of real-life tutor-student conversations is a considerable limitation. Synthetic dialogues, while useful for preliminary investigation, do not entirely capture the complexity and nuances of authentic tutor-student interactions. Second, the sample size of the dialogues used in this study may limit the generalizability of the findings. We used only 30 synthetic dialogues for this study, and increasing this number would likely improve the reliability and robustness of our findings.	 Third, the few-shot prompts we utilized were relatively simple and included a limited variety of examples. By integrating a wider range of nuanced examples, we might boost GPT-4's capability to match human graders' discernment of praise criteria that are more nuanced and socially sensitive.

\subsection{Implications for Future Work}
The present work sets a precedent for potential expansions. Firstly, we aim to address existing limitations by incorporating real-life dialogues, increasing the volume of chat logs, and enhancing the effectiveness of zero-shot and few-shot prompting methods. Secondly. the scope could be broadened by evaluating dialogues using a more comprehensive, high-level tutoring rubric. This would move away from focusing solely on specific tutoring skills such as delivering effective praise to students. As previously discussed, and recommended by the National Student Support Accelerator \cite{NSSA2021} for adoption by tutoring organizations, the holistic tutoring rubric could lay the groundwork for future efforts in crafting LLM prompts. These prompts could then provide timely feedback to tutors regarding their overall performance. Thirdly, apart from investigating the accuracy of GPT-4's performance, we could delve into other aspects, such as its reliability in synthesizing such feedback.

\section{Conclusion}

In this study, we assigned GPT-4 the task of identifying five distinct components of effective praise from synthetic tutor-student dialogues, according to past research determining criteria of effective praise. Our results suggest that GPT-4 performs moderately well in identifying two of these criteria: specific praise (which provides detail on what the student did well) and immediate praise (which is delivered right after the student's action). Conversely, GPT-4 had less success in recognizing instances of process-focused and sincere praise from the tutor. Overall, zero-shot and few-shot chain of thought prompting methods performed similarly. However, we anticipate enhancements to few-shot chain-of-thought prompting techniques, in particular, more nuanced and socially-responsive examples of sincere praise criteria will improve the performance of GPT-4 to detect praise closer to that of human graders.  \\

\noindent\textbf{Acknowledgments.} 
\\ This work is supporting with funding from the Richard King Mellon Foundation (Grant\#10851) and the Heinz Endowments (E6291). Any opinions, findings, and conclusions expressed in this material are those of the authors. Additionally, thanks Sorawit Saengkyongam, Can Udomcharoenchaikit, and Maim Hoque for contributing their thoughts on this research. 

\bibliographystyle{splncs04}
\bibliography{mybibliography}




% LNCS Homepage, \url{http://www.springer.com/lncs}. Last accessed 4
% \end{thebibliography}
\end{document}